\pdfoutput=1
\documentclass[11pt]{article}

\usepackage{acl}
\usepackage{scalerel}

\usepackage{times}
\usepackage{latexsym}

\usepackage[T1]{fontenc}
\usepackage[utf8]{inputenc}

\usepackage{microtype}
\usepackage{algorithm}
\usepackage{anyfontsize}
\usepackage{t1enc}
\usepackage{booktabs}
\usepackage{times}
\usepackage{amsmath}
\usepackage{algorithm}
\usepackage[noend]{algpseudocode}
\usepackage{diagbox}
\usepackage{latexsym}
\usepackage{enumitem}
\usepackage{siunitx}

\usepackage{helvet}  %Required
\usepackage{courier}  %Required
\usepackage{url}  %Required
\usepackage{graphicx}  %Required
\usepackage{courier}
\usepackage{amsmath}
\usepackage[normalem]{ulem}
\useunder{\uline}{\ul}{}
\usepackage{graphicx}
\usepackage{multirow}
\usepackage{mathbbol}
\usepackage{caption}
\usepackage{CJKutf8}
\usepackage{verbatim}
\usepackage{float}
\usepackage{booktabs}
\usepackage{array}
\usepackage{times}
\usepackage{csquotes}
\usepackage{latexsym}
\usepackage{tablefootnote}
\usepackage[normalem]{ulem}
\usepackage{todonotes}
\usepackage{subcaption}

\title{Zero-shot Sonnet Generation with Discourse-level Planning \\and Aesthetics Features} %by Training on Non-poetic Texts 

\author{Yufei Tian \\
  Computer Science Department, \\
  University of California, Los Angeles\\
  \texttt{yufeit@cs.ucla.edu} \\\And
  Nanyun Peng \\
  Computer Science Department, \\
  University of California, Los Angeles\\
  \texttt{violetpeng@cs.ucla.edu} \\}

\begin{document}
\maketitle
\begin{abstract}
Poetry generation, and creative language generation in general, usually suffers from the lack of large training data. %Moreover, \textit{mimicking} existing corpora in principle deviates from the goal of generating \textit{novel} contents. 
In this paper, we present a novel framework to generate sonnets that does not require training on poems. We design a hierarchical framework which plans the poem sketch before decoding. Specifically, a content planning module is trained on non-poetic texts to obtain discourse-level coherence; then a rhyme module generates rhyme words and a polishing module introduces imagery and similes for aesthetics purposes. Finally, we design a constrained decoding algorithm to impose the meter-and-rhyme constraint of the generated sonnets. Automatic and human evaluation show that our multi-stage approach without training on poem corpora generates more coherent, poetic, and creative sonnets than several strong baselines.\footnote{Our code and data are available at \url{https://github.com/PlusLabNLP/Sonnet-Gen}.}
\end{abstract}

\section{Introduction}
A sonnet is a fourteen-line poem with rigorous meter-and-rhyme constraints. In this paper, we aim at generating full-length sonnets that are logically and aesthetically coherent, without training on poetic texts. 

There are several challenges for this ambitious goal. First, there are limited number of sonnets available to train a fully supervised model. The only resource is a mere 3,355 sonnets collected by \citet{lau2018deep} in Project Gutenberg~\cite{Gutenberg}, one of the largest free online libraries for English literature.
While it is possible to train on related corpus such as general poems or English lyrics \cite{ghazvininejad2016generating}, such approaches are not applicable to many languages for which sizable poetry/lyrics data do not exist. Moreover, even if large-scale creative texts exist, learning from and mimicking existing corpora is \textit{not} creative by definition and is unlikely to result in novel content. 

\begin{figure}[t!]
\centering
\includegraphics[width=0.92\columnwidth]{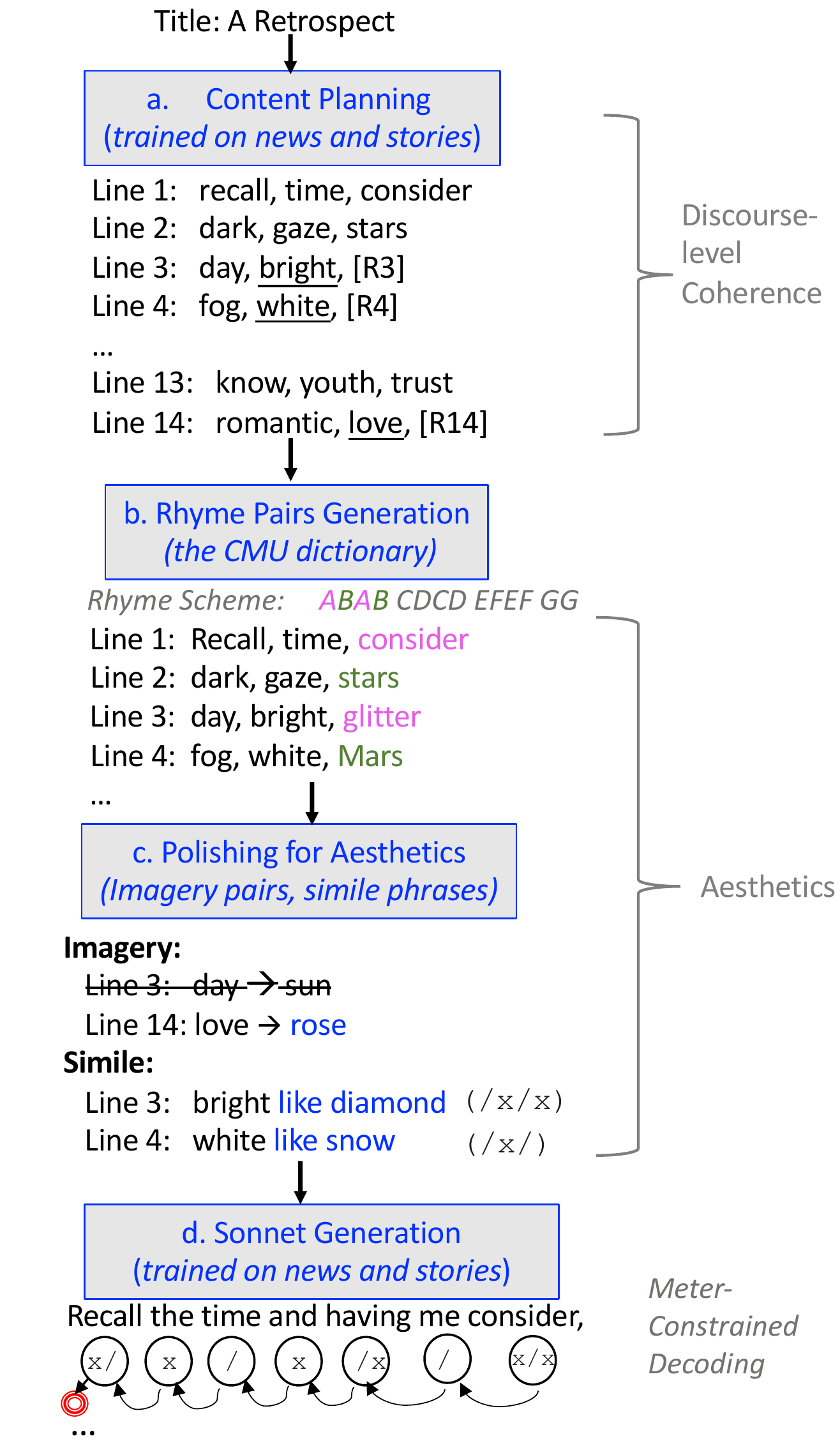} 
\vspace{-0.7em}
\caption{An overview of our approach. The content planning module generates keywords while maintaining discourse-level coherence. The second module form rhyming pairs and the polishing module enrich the imagination and add poetic flavor. (The keywords underlined in the first step have been polished.) Finally, we generate the sonnet with a meter-constrained decoding algorithm. Note that all four steps do not require poem/sonnet data.
}
\vspace{-1.7em}
\label{fig:illus1}
\end{figure}

Second, coherence remains a known issue among previous works on poetry generation. Existing works mainly focus on conforming to the format constraints (i.e., meter-and-rhyme), or generating a small stanza with a typical length of four \cite{lau2018deep,liu2019rhetorically,yi2020mixpoet}. For full-length sonnets, \citet{ghazvininejad2016generating} propose to use topical words as rhyme words to achieve topical relatedness, but the generated sonnets are not discourse-level coherent. They later generate discourse-level coherent English sonnets through French-English translation \citep{ghazvininejad2018neural}. Generating logically and aesthetically ordered poems without relying on content translation from other languages remains a challenge.

With all these in mind, we propose \ModelName{}, a \textbf{Ze}ro-shot \textbf{s}onne\textbf{t} \textit{generation model that does not require training on any poetic data}.
%\violet{Then, I suggest we name each of your 4 components for easy discussion. Also, you still haven't emphasize ``no need to train on poetic texts'', and I think that should be the most important thing you highlight}
Our framework, as is shown in Figure \ref{fig:illus1}, consists of four components: content planning, rhyme pairing, polishing for aesthetics, and final decoding. The first three steps provide salient points for the sketch of a sonnet. The last step is responsible for ``translating'' the sketch into well-formed sonnets.

To achieve zero-shot generation, the content planning and the final decoding components are both trained on a combination of news and story corpora. 
The trained planning module is aimed to generate several keywords for each sentence to equip the system with \textit{general world knowledge to construct a coherent text world}. 
However, the language used by poems is different from that of standard texts because it follows certain rhetorical rhythm and is full of vivid descriptions that appeals to readers' senses and imagination \cite{gibbs1994poetics}. To this end, in the polishing step we leverage external knowledge and incorporate two figurative speeches (i.e., simile and imagery) into the planned keywords to boost vividness and imagination. The rhyme and final decoding steps are designed to impose the meter-and-rhyme constraints.

While there are previous works on creative generation using the \textit{plan-and-write} paradigm~\cite{wang2016chinese, martin2018event,peng2018towards,yao2019plan,gao2019discrete,goldfarb2019plan},
 they all rely on training data from the target task domain (e.g., use story data to train storyline-planning). We on the other hand adopt content planning to disentangle the training from the decoding step to circumvent the shortage of training data for poetry generation. We summarize our contributions as follow:
\begin{itemize}[leftmargin=*]
\vspace{-0.3em}
    \item We propose \ModelName, a \textbf{Ze}ro-Shot \textbf{s}onne\textbf{t} generation framework, by disentangling training from decoding. Specifically, we first learn to predict context and rhyme words from news and story dataset, and then polish the predicted keywords to promote creativity. A constrained decoding algorithm is designed to impose the meter-and-rhyme constraints while  incorporating the keywords. 
\vspace{-0.3em}    \item We develop two novel evaluation metrics to measure the quality of the generated poems: automatic format checking and novelty evaluation (i.e., diversity and imageability).
 \vspace{-.3em}  \item Human evaluation shows that \ModelName{} generates more discourse-level coherent, poetic, creative, and emotion-evoking sonnets than baselines.
\end{itemize}

%\violet{I merged your original section 2 and section 3. I'll add a transitional paragraph here to give an overview about what you'll talk about in this section.}
\section{Background}
%In this section, we provide the background knowledge for automatic sonnet generation. Section \ref{sec:sonnet-struct} to \ref{sec:rhyme} 
In this section, we introduce the characteristics of sonnets in terms of structure, meter and rhyme. We then define important terminologies.
\subsection{The Structures of Sonnets}\label{sec:sonnet-struct}
 We aim to generate the two most representative sonnets: \textit{Shakespearean} and \textit{Petrarchan}. Sonnets make use of rhymes in a repeating pattern called \textbf{rhyme schemes} as shown in Table \ref{table:shakes-petra}. For example, when writing a Shakespearean sonnet, poets usually adopt the rhyme scheme of ABABCDCDEFEFGG. Although all sonnets have 14 lines, a Petrarchan sonnet consists of an 8-line stanza called an octave followed by a 6-line stanza called a sestet. On the other hand, a Shakespearean sonnet consists of three 4-line quatrains and a 2-line rhyming couplet which leaves the reader with a lasting impression. %This rhyming couplet signifies a succinct end to this poetic form, summarizing the meaning of the poem and leaving the reader with a lasting impression.

\begin{table}[h]
\small
\centering
\vspace{-.7em}
\begin{tabular}{@{\ \ }l@{\ \ }c@{\ \ }c@{\ \ }c@{\ \ }c@{\ \ }}
\toprule
\textbf{}                                                                 & \textbf{\begin{tabular}[c]{@{}l@{}}\# of\\ Lines\end{tabular}} & \textbf{\begin{tabular}[c]{@{}l@{}}Iambic \\ Penta\end{tabular}} & \textbf{Structure}                                                      & \textbf{\begin{tabular}[c]{@{}l@{}}Rhyme\\ Scheme\end{tabular}} \\ \midrule

\textbf{\begin{tabular}[c]{@{}l@{}}Shakespearean \\Sonnet\end{tabular}} & 14                                                             & Yes                                                                    & \begin{tabular}[c]{@{}l@{}}3 quatrain\\ 1 couplet\end{tabular} & \begin{tabular}[c]{@{}l@{}}ABAB\\ CDCD\\ EFEFGG\end{tabular} \\ \midrule

\textbf{\begin{tabular}[c]{@{}l@{}}Petrarchan  \\ Sonnet\end{tabular}}    & 14                                                             & Yes                                                                    & \begin{tabular}[c]{@{}l@{}}1 octave \\ 1 sestet\end{tabular}       & \begin{tabular}[c]{@{}l@{}}ABBA\\ ABBA\\ CDECDE\end{tabular} \\
\bottomrule
\end{tabular}
%\caption{The structure of a Shakespearean sonnet.}
\vspace{-.5em}
\caption{Comparison between a Shakespearean sonnet and a Petrarchan sonnet.}
\label{table:shakes-petra}
\vspace{-1.7em}
\end{table}

%\subsection{The Context of Sonnets} We all know that the language in poems is different from that in normal texts such as prose, making the appreciation of poems also different from others. Sonnets, as one representative type of poems, adopt descriptive and vivid language that often has an economical or condensed use of words \cite{sherman1893analytics}.

\subsection{Meter Constraints}\label{sec:meter-cons}
Most sonnet conform to iambic pentameter, a sequence of ten syllables alternating between unstressed (\texttt{x} or \texttt{da}) and stressed syllables (\texttt{/} or \texttt{DUM}). Strictly speaking, each line reads with the rhythm (\texttt{da-DUM})$^5$, which enhances the tone for the poem and operates like an echo. In reality, there are many rhythmic variations. For example, the first foot is often reversed to sound more assertive, and can be written as (\texttt{DUM-da} * (\texttt{da-DUM})$^4$). Another departure from the standard ten-syllable pattern is to append an addition unstressed syllable to the end, forming feminine rhymes which can be written as ((\texttt{da-DUM})$^5$*\texttt{da}).

\subsection{Rhyme Words, Couplets and Patterns}\label{sec:rhyme}
A pair of \textbf{rhyme words} consists of two words that have the same or similar ending sound. A \textbf{rhyming couplet} is a pair of rhymed lines. For example, Line 1\&3, 2\&4 in Figure \ref{fig:illus1} are two pairs of rhyming couplets. From the CMU pronunciation dictionary %\footnote{https://github.com/aparrish/pronouncingpy}
\cite{CMUdict}, we know that ``fall" and ``thaw" in Figure \ref{fig:illus1} are \textit{strict} rhyming pairs because they have exactly the same phonetic endings: \texttt{"\textbf{AO} L"}. ``Leaves" (\texttt{"\textbf{IY} V Z"}) and ``trees" (\texttt{"\textbf{IY} Z"}) are \textit{slant} rhymes, because they have the same stressed vowels, while the ending consonants are similar but not identical.

\subsection{Terminology}\label{sec:term}
%Before we come to the model architecture, 
We formally define the following terms:
\begin{itemize}[leftmargin=*]
 \vspace{-0.7em}    \item Keywords $\mathcal{K}$: content words and rhyme words combined. They contain main ideas of a poem and define the rhyming pattern.
\vspace{-0.7em}    \item Content words $\mathcal{C}$: keywords that do not appear in the end of each line. We target at predicting 2 context words per line, $C_{i1}$ and $C_{i2}$.
 \vspace{-0.7em}   \item Rhyme words $\mathcal{R}$: words in the end of each line. For example, in a Shakespearean sonnet with the rhyme scheme ABABCDCDEFEFGG, there are seven pairs of rhyme words: $R_1R_3$, $R_2R_4$, %$R_5R_7$, $R_6R_8$, $R_9R_{11}$, $R_{10}R_{12}$,
 ..., and $R_{13}R_{14}$.
\vspace{-0.7em}    \item Initial rhyming lines $\mathcal{I}_{Init}$: index of the lines that the first rhyme word in a rhyming couplet appears (e.g.,  $\mathcal{I}_{Init}$ = [1, 2, 5, 6, 9, 10, 13] for a Shakesperean sonnet and $\mathcal{I}_{Init}$ = [1, 2, 9, 10, 11] for a Petrarchan sonnet).
 \vspace{-0.7em}   \item Sketch: The sketch of a poem contains three aspects: 1) content words that cover the key concepts or main ideas, 2) the rhyme words to appear at the end of each line, and 3) the modification of keywords for aesthetics.
\end{itemize}
\section{Approach} \label{sec:method}
\paragraph{Overview}
As is shown in Figure \ref{fig:illus1}, our sonnet generation model can be divided into four steps. At step a, we train a title-to-outline module by finetuning T5 \cite{raffel2019exploring} on keywords extracted from news reports and stories. During inference time, we generate a fourteen-line sonnet sketch that contain those content words $\mathcal{C}$ (Section \ref{subsec:a}).
At step b, we aim at forming the correct rhyming pairs. We first select the initial rhyme words from $\mathcal{C}_i$ for $i \in \mathcal{I}_{Init}$, and then generate the remaining rhyme words (i.e., for  $i \in \overline{\mathcal{I}_{Init}}$) by forcing the decoder to sample from a vocabulary pool that contains strict and slant rhyme words (Section \ref{sec:gen-rhyme}). 
At step c, we infuse imagery and simile as two figurative devices to   $\mathcal{C}$ (Section \ref{subsec:c}). 
In the last step, we leverage a fine-tuned language model with constrained decoding algorithm to impose the meter-and-rhyme constraints (Section \ref{sec:sonnet-gen}).% We describe each component subsequently. 

\begin{figure}[t!]
\centering
\includegraphics[width=1.0\columnwidth]{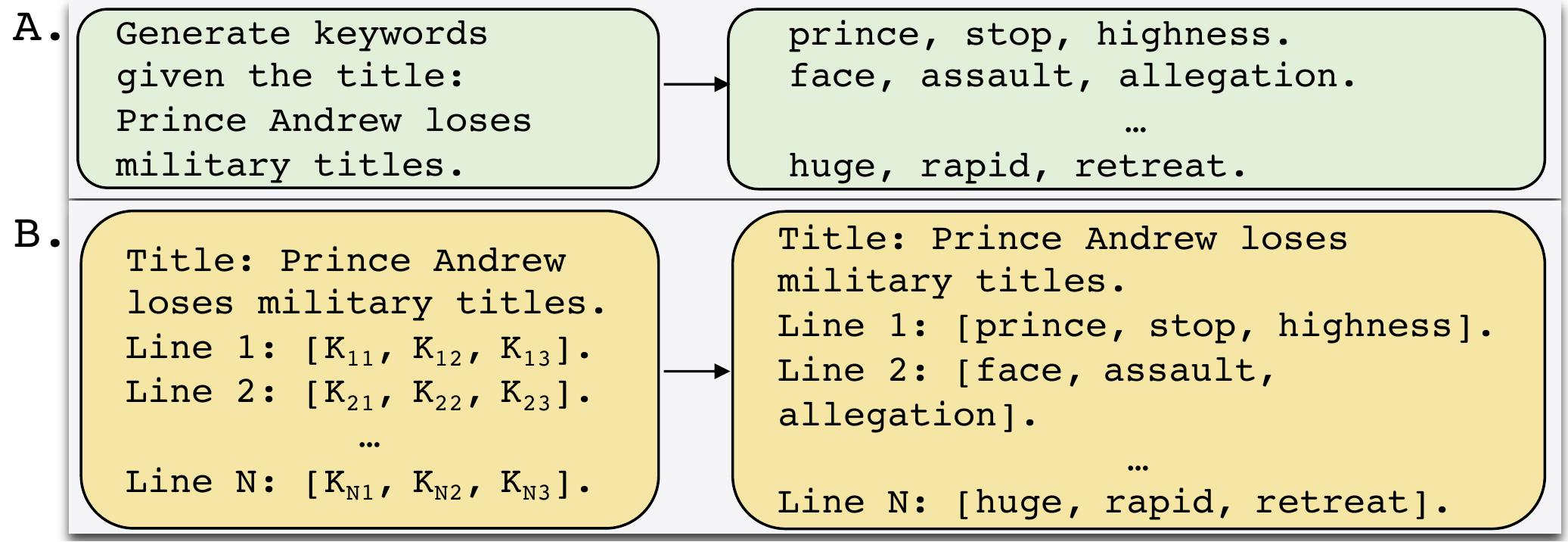} 
\caption{A comparison diagram of two input-output formats to train the first module. While format \texttt{A} is most straight-forward, there is no control over the output structure. Therefore, we purposefully design the prompt shown in format \texttt{B} to control the number of keywords and the number of lines to be generated. $\mathcal{K}_{ij}$ represents the mask tokens at the $i$-th sentence.}
\label{fig:T5}
\vspace{-1.3em}
\end{figure}

\subsection{Content Planning}\label{subsec:a}
For each piece of news or stories, we train a title-to-keywords framework that predicts the outline. To this end, we first extract three most salient words per line using the RAKE \cite{rose2010automatic} algorithm, which is a domain-independent keyword extraction technique.

\paragraph{Controllable Text Formatting} We then leverage the task adaptability of the pretrained T5 \cite{raffel2019exploring} to predict the keywords of the whole body. As a unified framework that treats every text processing task as a “text-to-text” problem, T5 can be easily adapted to our task as shown in Figure \ref{fig:T5}.A, where the input is an instruction to generate the sketch given the title, and the outputs are multiple keywords for each line.
However, we need a mechanism to specify the number of lines and keywords to be generated, since we train on prosaic texts with varying formats but infer only on the 14-line sonnets.

To solve this problem and gain control over the poem structures, we format the input and output as shown in Figure \ref{fig:T5}.B. Specifically, we use [MASK] tokens as placeholders for the keywords. Now that one [MASK] token on the input side corresponds to exactly one word on the output side, we are able to specify the number of lines and keywords during the inference time.

\subsection{Generating Rhyme Words} \label{sec:gen-rhyme}
Our title-to-outline model is trained to generate keywords, regardless of the rhyme constraints. In this section, we describe the procedure to generate rhyme pairs. 
Specifically, we force the model to generate a 14-line outline, with two or three content words for each line depending on whether the line is an initial rhyming line:
\begin{equation}
%\vspace{-0.9em}
    \text {Keywords}_i
=\left\{\begin{array}{ll}
{\left[K_{i 1}, K_{i 2}, K_{i 3}\right], \text { if $i$ in $\mathcal{I}_{Init}$ }} \\
{\left[K_{i 1,}K_{i2}\right], \text {otherwise.}}
\end{array}\right.
\label{eqa:masks}
\end{equation}

\noindent where $K_{i j}$ represents the $j$-th keyword in the $i$-th line. Among the three keywords in the initial rhyming lines, we select the last word as the initial rhyme word. 
\begin{figure}[t]
\centering
\includegraphics[width=0.85\columnwidth]{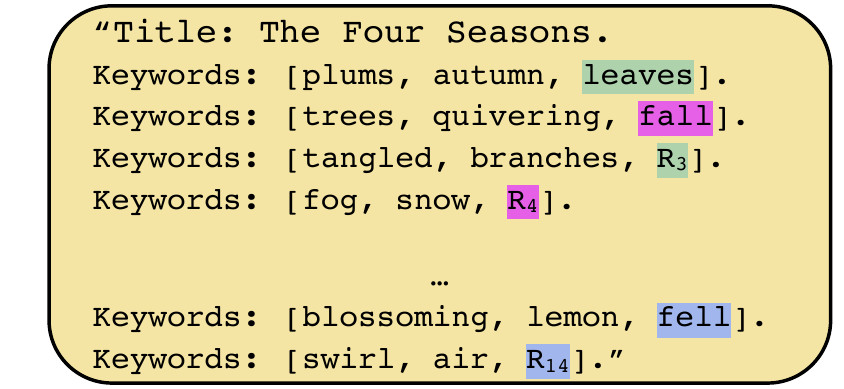} 
\caption{An example input to query the remaining rhyme words during the inference time. Rhyme words in the same background color form a rhyming pair.}
\label{fig:mask2}
\vspace{-1.7em}
\end{figure}     

\paragraph{Rhyme Pairs Generation} Given the initial rhyme words, we then retrieve all the possible rhyme words $\mathbb{R}$ based on their phonetics information from the CMU prounounciation dictionary \cite{CMUdict}. This include strict rhymes and slant rhymes. For instance, in Figure \ref{fig:mask2}, the retrieved rhyme word candidates $\mathbb{R}$ for `leaves' are [`achieves', `believes', `Steves', `trees',  ...]. The probability distribution for generating the rhyme word $w_R$ from the candidate list $\mathbb{R}$ is modified as:
\begin{equation}
%\vspace{-0.9em}
P^{\prime}(w_R)=\left\{\begin{array}{cc}
\frac{p(w_R\mid \text {context})}{\sum_{x \in \mathbb{R}} p\left(x \mid \text{context} \right)} &, \text{if } w_R \in \mathbb{R} \\
0 &, \text {otherwise.}
\end{array}\right.
\end{equation}

\noindent where $p(w_R | \cdot) $ is the original word probability yielded by the title-to-outline decoder.

\subsection{Polishing Context Words for Aesthetics} \label{subsec:c}
Now, we have the generated context words and rhyme words that are discourse-level coherent yet less vivid. To this end, we use external knowledge to incorporate two figurative devices into the planned keywords: imagery and simile. 

\paragraph{Imagery} We leverage the <symbol, imagery> pairs (e.g., <love, rose>) in the ConceptNet knowledge base \cite{liu2004conceptnet} and finetune a imagery generation model from a pretrained model called COMmonsEnse Transformer \cite{bosselut2019comet} (COMeT). It is trained on imagery pairs to generate the imagery word given the symbolism word as input. At inference time, we randomly sample multiple nouns from the sketch to predict their imageries, and only make replacement for the two most confident generations. For example in Figure \ref{fig:illus1}, both <day, sun> and <love, rose> are generated, yet we only replace `love' with `rose', because the probability of generating the latter pair is much higher than the former pair.

\paragraph{Simile} A simile phrase consists of two parts: the adjective and the figurative vehicle. For example, `sudden like a flash' is a simile phrase where `a flash' is the figurative vehicle of `sudden'. We leverage the simile generation model by  \citet{chakrabarty2020generating} as an off-the-shelf tool\footnote{https://github.com/tuhinjubcse/SimileGeneration-EMNLP2020} to generate simile vehicles from adjectives to extend the sketch keywords. %and calculate the probability.
At inference time, we randomly sample multiple adjectives from the sketch to predict their figurative vehicles, and only keep the most confident ones. In addition, we also make sure the generated simile phrase conforms to the iambic-meter constraint. For example in Figure \ref{fig:illus1}, the phrase `bright like diamond' (/x/x) follows the iambic meter, whereas another phrase such as `shining like diamond' (/xx/x) will be disregarded. 

\subsection{Sketch to Sonnet Generation}\label{sec:sonnet-gen}
% We fine-tune GPT-Neo-2.7B on the same combination of news and stories data as a language model to generate the sonnet.%\violet{I think at some point, maybe early on, even in the intro, we should simply say what's the dataset we use -- I feel that's easier... keep mentioning ``the standard text'' just feel mouthful.}
\begin{algorithm}[t]
\small
\caption{Gen Valid Tokens}
\begin{algorithmic}[1]

\Function{Gen}{ $gen_t$, $stress_t$}\\%\Comment{Where A - prompt}\\
    \textbf{Parameter}: Int - $t$ \Comment\textit{\textcolor{gray}{current time step}} \\
    \textbf{Parameter}: Int - $N$ \Comment\textit{\textcolor{gray}{num of return samples}} \\
    \textbf{Parameter}: List - $CW$ \Comment\textit{\textcolor{gray}{context words yet to include}} \\
      \textbf{Input: }List of strings - $gen_t$, $stress_t$ \Comment\textit{\textcolor{gray}{generated beams at time step $t$ and corresponding 0/1 stress series}}\\ 
      
      \textbf{Output: }{List of strings - $gen_{t+1}$,  $stress_{t+1}$}%\Comment\textit{\textcolor{gray}{generated beams at time step $t+1$ and the corresponding 0/1 stress series}}
      \\

    \textbf{Initialize} $gen_{t+1}$, $stress_{t+1}$ to empty \\

\textbf{for} {$gen$, $stress$ in zip($gen_t$, $stress_t$)} \textbf{do}\\ \Comment\textit{\textcolor{gray}{repeat topk sampling N times and return all generations}}
    
        \State $tokens$ $=$ generate\_next($gen$, $N$).to\_set()
        \For{$c$ in $CW$}
        \If{$c$ not in tokens}
        \State $tokens$.append($c$)
        \EndIf
        \EndFor
        
        %\textbf{For} {$t$ \textbf{in} SET($tokens$)} \textbf{do}
        \For{$t$ in $tokens$}
        \Comment\textit{\textcolor{gray}{check for meter constraints}}
                \If {satisfy($t$, $stress$)}
                \State update $gen_{t+1}$, $stress_{t+1}$, $CW$
                \Else
                \State continue
                \EndIf
        \EndFor

        \Return{$gen_{t+1}$,  $stress_{t+1}$}
       \Comment{call recursively until 10 or 11 syllables are generated and disregard the metric line unless all three keywords are incorporated. }
       \EndFunction

\end{algorithmic}
\label{algo:1}
\vspace*{-1.0mm}
\end{algorithm}
In order to write fluent and poetic languages that meet the meter-and-rhyme constraints, we make the following adaptations. First, generating the full sonnet requires more powerful pretrained model than generating the outlines. Therefore, we fine-tune GPT-Neo-2.7B on the same combination of news and stories data as a language model to generate the sonnet. Second, to effectively incorporate the rhyme words at the end of each line, we follow previous methods \cite{ghazvininejad2016generating, van2020automatic} and generate the whole sonnet line-by-line \textit{in reverse}, starting from the final rhyme word to the first word. That is to say, our language model is finetuned to generate from right to left to better enforce rhyming. 
Third, we include the sketch in the prompt, so that the decoder will learn to give higher probability for these keywords. We then use lexically constrained decoding similar to that of Grid Beam Search \citep{hokamp2017lexically} to incorporate the keywords. %In this way the model is more likely to incorporate the keywords in the completion. 
In addition, we also include the previously generated lines in the prompt to generate the next line in a sonnet to promote discourse-level coherence. 
A simile phrase in the sketch is considered fixed that cannot be modified. Namely, we force to generate the whole phrase when the first word in the phrase is decoded. Lastly, we modifies the beam search algorithm to impose the meter-and-rhyme constraint. Algorithm \ref{algo:1} displays the skeleton of our decoding strategy. At each decoding step, we apply rhythm control, so that only those tokens that satisfy the iambic-pentameter and its two variations (listed in Section \ref{sec:meter-cons}) are kept in the beams. We recursively generate the next token until 10 or 11 syllables are generated and make up a metric line where all the context words are incorporated.

%\vspace*{-1mm}
\section{Experimental Setup}
\subsection{Dataset}
Our approach does not require poem data. 
The training dataset for the content planing module and the decoding module is a combination of 4,500 CNN news summary \cite{HermannKGEKSB15} and 16,000 short stories crawled from Reddit.\footnote{https://www.reddit.com/r/shortscarystories/} We remove those articles that contain conversations, urls, or are too long (>50 lines) or too short (<8 lines). During decoding, we generate sonnets using top-k sampling and set no\_repeat\_ngram\_size to 3 to promote creativity and avoid repetition.

We finetune the pretrained T5 for 10 epochs for the ``content planning'' component,  and finetune GPT-Neo-2.7B for 6 epochs for the decoding component. We use one Nvidia A100 40GB GPU. The average training time is 5$\sim$10 hours for each experiment.

\subsection{Baselines}
\paragraph{Hafez} A program that is trained on lyrics data and generates sonnets on a user-supplied topic~\cite{ghazvininejad2018neural}. It combines RNNs with a finite state automata to meet the meter and rhyme constraints. Hafez is the state-of-the-art model that generates full-length sonnets but it does not train on standard, non-poetic texts. 

\paragraph{Few-shot GPT-3}
We utilize the most capable model in the GPT-3 family \cite{brown2020language}, \textit{GPT3-davinci}\footnote{https://beta.openai.com/docs/engine}, as a strong baseline to follow instructions and generate sonnets. In the prompt, we provide two examples of standard sonnets and then instruct the model to generate a sonnet given the title. We force the output to be exactly 14 lines.

\paragraph{Ablations of our own model} To test the effectiveness of our sketch-before-writing mechanism, we also compare variations of our own model:
\paragraph{\ProsaicName} An stronger version of \textit{nmf} \cite{van2020automatic}, the first (and only) model to generate rhyming verses from prosaic texts. Topical and rhyme consistency are achieved by modifying the word probability of rhyme and topical words. For fair comparison, we replace the original vanilla encoder-decoder with GPT2 that \ModelName{} is finetuned on, and force the output to be 14 lines. Model comparison between \ProsaicName{} and \ModelName{} serves as ablations of the keyword-planning component (versus end-to-end generation).

\paragraph{\VTwoName} The model consisting of step a, c, and d as illustrated in Figure \ref{fig:illus1}, but without the polishing the sketch for figurative devices.  Our full model consisting of 4 modules is called \ModelName.

%\paragraph{Summary}

\subsection{Decoding Strategy}
For decoding, we generate sonnets from our models using a top-k random sampling
scheme where k is set to 50. At each time step, the GPT2 model generates subwords instead of complete words. In order to impose the meter and rhyme constraints while decoding for each word, we ask the language model to continue to generate until a complete word is generated as indicated by special space token `Ġ'.  To avoid repetition and encourage creativity, we set no\_repeat\_ngram\_size to 3 and use a softmax temperature of 0.85.

\subsection{Automatic Evaluation}
It is difficult and thus uncommon to automatically evaluate the quality of poems. For example, \citet{ghazvininejad2016generating} and \citet{van2020automatic} exclude automatic evaluation, with the later stating ``Automatic evaluation measures that compute the overlap
of system output with gold reference texts such as
BLEU or ROUGE are of little use when it comes to creative language generation." In addition,  \citet{yang2020ubar} show current metrics have very low correlation with human. Hence, we propose to evaluate the generated poems in two novel aspects: format and novelty.
\paragraph{Format Checking} For rhyme checking, we count the percentage of rhyme pairs that belong to strict or slant rhymes. For meter checking, we consider the following most common scenarios mentioned in Section \ref{sec:meter-cons}: the standard Iambic Pentameter; the first foot reversed; and
a feminine rhyme.
In all scenarios, words that are monosyllables can serve as both stressed and unstressed syllables. For a looser standard, we also calculate the percentage of valid lines that contain either 10 or 11 syllables.

\paragraph{Novelty}
 We follow the settings in exsiting works \citet{yi2018automatic, yi2020mixpoet} and calculate the Distinct-2 scores \cite{li2015diversity} to measure the diversity of generated poems.
 Besides, imagery is another important feature of poems as pointed out by linguistic studies \citet{kao2012computational, silk2006interaction}. 
 Here, we calculate \textit{Imageability} score to assess how well a poem invokes mental pictures of concrete objects.  Specifically, we extracted the features from the resource by \citet{tsvetkov2014metaphor}, who use a supervised learning algorithm to calculate the imageability ratings of 150,114 terms. For each poem, we average the ratings of all its words after removing the stop words.

\subsection{Human Expert Judgement}
% Currently available automatic metrics cannot fully reflect the quality of poems. Hence, we also conduct human-based evaluation.
 Considering the expertise required to appreciate sonnets, we recruit 6 professionals that hold a bachelor's degree in English literature or related majors as domain experts to annotate the generated sonnets. We provide detailed instructions and ask them to evaluate the each poem on a scale from 1 (not at all) to 5 (very) on the following criteria: 
 \textbf{1) Discourse Coherence}: whether the sonnet is well organized, with the sentences smoothly connected and flow together logically and aesthetically,
 \textbf{2) Originality/Creativity}: the usage of original ideas in the poem, including imagination, rhetorical devices, etc., 
 \textbf{3) Poetic in language}: how well the poem adopts descriptive and vivid language that often has an economical or condensed usage, 
 \textbf{4) Emotion Evoking}: if the poem is emotionally abundant and make the readers emphasize with the writer.
 At last, we ask the annotators to judge if the sonnet is written by a poet with \textit{serious} goals to write a poem. In total, we evaluate 50 sonnets for each baseline and the gold standard (human) model. Each sonnet is rated by three professionals. 
 
 The average inter-annotator agreement (IAA) in terms of Pearson correlation is 0.61 with p-value <0.01, meaning that our collected ratings are highly reliable. We also conduct paired t-test for significance testing. The difference between our best performing model and the best baseline is significant.  Considering the expertise required, human evaluators are paid \$25 per hour.
 
\begin{table}[t!]
\small
\centering
\begin{tabular}{@{}l@{ }|c@{ }|@{ }c@{ }|@{ }c@{ }|c@{ }|@{ }c@{ }}
\toprule
    \multirow{2}{*}{Model Name}    & \multicolumn{3}{@{}c@{}|}{\textbf{Format Checking}}                                            & \multicolumn{2}{@{}c@{}}{\textbf{Novelty}}                              \\ \cmidrule(l){2-6} 
 & Rhyme                        & Meter                       & Syllable                                         & Dist-2                       & Img \\ \midrule
Hafez                                            & { 98.3\%} & { 76.8\%} & 95.7\%                                                & 84.8                        & 0.44       \\
Fewshot GPT-3                                            & 14.0\%                        & 17.6\%                        & 30.9\%                                              & 85.3                        & 0.48       \\ \midrule
\ProsaicName                                         & { \ul 100\%}  & { 10.1\% }                       & {19.0\%}                         & {84.9}                        & 0.46       \\ 
\VTwoName                                         & { \ul100\%}  & {\ul 77.7\%} & {\ul 98.6\%}  & {\ul86.6} & {0.49}       \\
\ModelName                                        & {\ul100\%}  & 75.6\%                        & 98.4\%                         & {\ul 86.6} & {\ul0.51}       \\  \midrule %\hline
\textbf{Human}                                            & {94.6\%}               & {70.7\%}               & {81.8\%}                            & {87.4}               & 0.52  \\ \bottomrule
\end{tabular}
\caption{Automatic evaluation results for rhyme, meter, syllable checking, distinct scores, and imageability (Img in the table). Best machine scores are underlined.}
\label{table:auto-results}
\vspace{-1em}
\end{table} 

\section{Results and Analysis}
\subsection{Results of Automatic Evaluation}
Table \ref{table:auto-results} summarizes the format checking and novelty scores of our model compared to the baselines. We can see that human poets tend to incorporate more variations and do not strictly follow the meter and rhyme constraints, which computers are good at. GPT-3 fails to learn the sonnet formats through massive pretraining and few-shot learning despite its gigantic size. \ProsaicName{} falls short of meter-checking because is only trained to generate rhyming verses. Since we utilize the the phonetics information provided in the CMU dictionary, \ModelName{} achieves 100\% success in rhyme words pairing. As for novelty, \ModelName{} generates most diversely and is best at that arousing mental pictures of concrete objects among machines.

\begin{table}[t!]
\small
\centering
\renewcommand{\tabcolsep}{1.2mm}
\begin{tabular}{l|c|c|c|c|c}
\toprule
         %& \multicolumn{1}{l|}{CH} & \multicolumn{1}{l|}{CR} & \multicolumn{1}{l|}{PL} & \multicolumn{1}{l|}{EE} & \multicolumn{1}{l|}{TP} & \multicolumn{1}{l}{WH} \\ \midrule

        & \multicolumn{1}{c|}{\begin{tabular}[c]{@{}l@{}}DC\end{tabular}} & \multicolumn{1}{c|}{\begin{tabular}[c]{@{}l@{}}O\end{tabular}} & \multicolumn{1}{c|}{\begin{tabular}[c]{@{}l@{}}P\end{tabular}} & \multicolumn{1}{c|}{\begin{tabular}[c]{@{}l@{}}E\end{tabular}} &  \multicolumn{1}{c}{\begin{tabular}[c]{@{}l@{}}WH\end{tabular}} \\ \midrule
Hafez         & 3.09          & 3.01          & 3.05          & 2.95          & 41.3\%          \\
Few-shot GPT3 & 3.43          & 3.10          & 2.86          & 3.11          & 52.7\%          \\ \midrule
\ProsaicName      & 3.25          & 2.95          & 2.97          & 2.98          & 46.0\%          \\ 
\VTwoName     & {\ul 3.57}*    & 3.25          & 3.35          & 3.13          & 58.7\%          \\
\ModelName    & 3.52          & {\ul 3.41}*    & {\ul 3.66}* & {\ul 3.22}*    & {\ul 62.0\%}*     \\ \midrule
Human         & \textbf{3.82} & \textbf{3.54} & {\textbf{ 3.68}}    & \textbf{3.56} & \textbf{83.3\%}                                         \\ \bottomrule
\end{tabular}
\caption{Expert ratings on several criteria to assess sonnet quality: discourse-level coherence (DC), originality/creativity (O), poeticness in language (P), emotion evoking (E), and written by human (WH). We show average scores with 1 denoting the worst and 5 the best. We boldface/underline the best/second best scores. $*$ denotes that paired t-test shows that our model variations (\VTwoName, and \ModelName) outperform the best baseline in all aspects with statistical significance (p-value < 0.05).}
\label{table:human-results}
\vspace{-1.7em}
\end{table}

\subsection{Results of Human Evaluation}
Table \ref{table:human-results} presents the performance of the aforementioned evaluation criteria: coherence, originality, poeticness, and emotion-evoking. Our models (\VTwoName, and \ModelName) outperform the baselines in all aspects by a large margin. 

\paragraph{Comparison between our own models.}Compared with \ProsaicName{} which also generates poems from non-poetic texts, our models
generates more coherent sonnets with great statistical significance (p-value < 0.01), showing the superiority of explicit sketch planning over generating from scratch (i.e., end-to-end generation). 

\VTwoName{} generates more coherently than \ModelName{} (p-value < 0.10). However, \ModelName{} achieves high scores in originality, poeticness by a large margin (+0.2). Hence, we still consider it as our best model. It is also noteworthy that \ModelName{} is the most \textit{emotion-evoking} system among all machines even though we do not have explicit sentiment control. Poem theories have shown that emotion appeals lie in the following aspects: the general topic, the word choice, vivid descriptions, figurative language, insights and experience \cite{scheub2002poem}. We posit that aesthetic features in the \ModelName{} arouse emotion appeals.

\paragraph{Analysis for high poeticness.} \ModelName{} is on par with humans in terms of poeticness score, meaning that our models generate highly descriptive, vivid, and condensed text. With manual examination, we attribute such high poeticness to three aspects. First, the imagery and similes clearly represents traits of poems. Second, in keyword-planning we ensure that at least three concepts will be presented per line, and thus the generation module naturally become economical in word usage to include all the information. Lastly, with the constraint decoding algorithm to insert keywords, we inevitably become less natural (e.g., miss conjunctions and auxiliary verbs). While this can be a drawback in other generation tasks, the occasional omission of such auxiliary words is just opportune for sonnets, and adds to the flavor of a poem. The examples in table \ref{table:four-seasons} helps demonstrate these points.

\begin{table*}[t!]
\small
\centering
\begin{tabular}{|@{ }l@{}|l@{ }|l@{ }|l@{}|}
\toprule
\multicolumn{1}{|@{}c@{}}{\textbf{Hafez}}               &  \multicolumn{1}{|@{}c@{}}{Score}               & \multicolumn{1}{|@{}c@{}}{\textbf{\ProsaicName}}    &  \multicolumn{1}{|@{}c@{}|}{Score}                                                                                                    \\ \midrule
\fontsize{9}{8}\selectfont {
\begin{tabular}[c]{@{}l@{}}
Delightful hours over long \underline{ago},\\ 
Succeed to hang the little hook and \underline{light}.\\ 
The darker still and more extended \underline{snow},\\ 
The fresh invention bore the better \underline{night}.\\ \\ 
Another party started getting \underline{heavy},\\ 
And never had a little bit of \underline{summer}.\\ 
And got a trigger on the finger \underline{really},\\ 
Or something going by the name of \underline{winter}.\\ \\ 
Retreat the colors and fun in \underline{afternoon}\\ 
And bought the tickets loved a royal \underline{song}\\ 
A sacred person join and ride in \underline{moon}\\ 
Away the author go and winding \underline{long},\\ \\ 
 Oh maybe get a little out of \underline{sight},\\ 
 And wounded victim suffered from the \underline{site}.
\end{tabular} 
} &

\fontsize{9}{8}\selectfont {\begin{tabular}[c]{@{}l@{}}\\
C: 2.7\\ \\ \\ \\
O: 3.0 \\ \\ \\ \\
P: 3.0\\ \\ \\ \\
E: 3.0\\ \\ \\

\end{tabular}} &

\fontsize{9}{8}\selectfont {\begin{tabular}[c]{@{}l@{}}
The leaves fall and the air is \underline{cooler}, \\
Were we known with seasons? Spring is \underline{coming},\\
Of course there isn't that much \underline{humor},\\
And it is now \underline{blooming}.\\\\

Even when everyone else’s \underline{warm},\\
Someone would want to know what we could do in \underline{may},\\
else we didn’t known – It was the first \underline{storm},\\
It’s how they used to do every other \underline{day}.\\\\

What aren't you looking up Right behind us, no \underline{clouds},\\
Just know no one's will ever see the clouds or \underline{thunder},\\
And nobody other than the \underline{crowds},\\
Before now, we had just started to \underline{wonder}.\\\\

We all aren’t sure what we can \underline{keep},\\
they couldn't see me, like they were about to \underline{reap}.\\
\end{tabular}} &
\fontsize{9}{8}\selectfont {\begin{tabular}[c]{@{}l@{}}
C: 3.3\\ \\ \\ \\
O: 2.7 \\ \\ \\ \\
P: 3.0\\ \\ \\ \\
E: 3.3\\

\end{tabular}}

\\\toprule

\multicolumn{1}{|c|}{\textbf{\VTwoName}}  &  \multicolumn{1}{c|}{Score}&
\multicolumn{1}{c|}{\ModelName}  &  \multicolumn{1}{c|}{Score}\\ \midrule

\fontsize{9}{8}\selectfont {
\begin{tabular}[c]{@{}l@{}}
\underline{Falling} \underline{snow}, I feel a little \underline{cooler},\\
Is a \underline{sign} of stormy \underline{winter} \underline{coming},\\
Meet the \underline{honest} friends and \underline{gather} \underline{humor},\\
\underline{Happy} \underline{spring} and just admire the \underline{blooming}.\\\\

The \underline{heat} and ocean \underline{air} becoming \underline{warm},\\
\underline{Little} \underline{birds} and beautiful songs in \underline{May},\\
\underline{Leaves} and \underline{flowers} probably just the \underline{storm},\\
Among the lunar \underline{moons} and \underline{summer} \underline{day}.\\\\

\underline{Sudden} \underline{rain} and downpour from the \underline{thunder},\\
And \underline{summer} always \underline{fill} hotels with \underline{crowds},\\
Take a \underline{shower} and give the \underline{spring} a \underline{wonder},\\
Watch the \underline{blue} \underline{sky} and far behind the \underline{clouds}.\\\\

In \underline{months} the \underline{future} vegetables \underline{reap},\\
The \underline{years} and \underline{seasons} never really \underline{keep}.\\

\end{tabular} 
} &

\fontsize{9}{8}\selectfont {\begin{tabular}[c]{@{}l@{}}
C: 3.7\\ \\ \\ \\
O: 3.3 \\ \\ \\ \\
P: 3.3\\ \\ \\ \\
E: 3.0\\ 

\end{tabular}} &

\fontsize{9}{8}\selectfont {\begin{tabular}[c]{@{}l@{}}
The \underline{snow} is \underline{falling}, whether getting \underline{cooler},\\
Whispers {\color{blue}\underline{rumors}} of a \underline{winter} \underline{coming},\\
\underline{Gather} \underline{honest}, peace and love and \underline{humor},\\
{\color{blue}\underline{Blossom}} of the \underline{happy} season \underline{blooming}.\\\\

Wind gets \underline{heat} and the \underline{air} becoming \underline{warm},\\
The \underline{birds} enjoy the \underline{little} lovely \underline{may},\\
Beneath the \underline{leaves}, \underline{flowers} survive the \underline{storm},\\
The \underline{moon} is shining on a \underline{summer} \underline{day}.\\\\

{\color{blue}\underline{Sudden like a flash}} comes \underline{rain} with \underline{thunder},\\
The \underline{summer} vibes \underline{fill} the running \underline{crowds},\\
Because of \underline{shower}, \underline{spring} became a \underline{wonder},\\
The \underline{sky} is high and {\color{blue}\underline{blue like sea}} with \underline{clouds}.\\\\

The coming \underline{months} are watching \underline{future} \underline{reap},\\
Those \underline{years} and \underline{seasons} bring us all to \underline{keep}.\\
\end{tabular}}&

\fontsize{9}{8}\selectfont {\begin{tabular}[c]{@{}l@{}}
C: 3.7\\ \\ \\ \\
O: 4.0 \\ \\ \\ \\
P: 4.0\\ \\ \\ \\
E: 3.3\\ 

\end{tabular}}

\\ \bottomrule
\end{tabular}
\caption{An example of the generated sonnets from four systems with the same title: ``The Four Seasons". The scores are average numbers of three human ratings on the following criteria: coherence (C), originality (O), poetic in language (P), and emotion evokingness (E). We underline the planed keywords and highlight the figurative languages in blue.}
\label{table:four-seasons}
\vspace{-1.7em}
\end{table*}

\section{Qualitative Analysis}
\subsection{Case Study}
We conduct case study to better understand the advantages of our model over the baselines. Table \ref{table:four-seasons} lists the generated sonnets by Hafez, \ProsaicName{} and \ModelName{} given the same title: ``The Four Seasons".

\paragraph{Problems with the Baselines} Hafez chooses words that are related to the title as rhyme words.
However, topically related rhyme words are not sufficient for overall coherence. While it is locally understandable, the sonnet generated by Hafez is divergent and disconnected when sentences are put together. On the other hand, \ProsaicName{} mimics the rhyme and topical properties of poems, but still generate highly prosaic and colloquial sentences that are not poetic at all.  

\paragraph{Advantages of Our Model} Thanks to content planning, \VTwoName{} generates a well-organized sonnet that describes the four seasons from winter to autumn in a logical order. Despite minor grammar errors, the full model \ModelName{} benefits from vivid descriptions and natural imagery such as `whispers rumors of a winter coming', `blossom of the season', and `sudden like a flash'.

\subsection{Impact of Keywords}
\begin{figure}[t!]
\centering
\includegraphics[width=0.75\columnwidth]{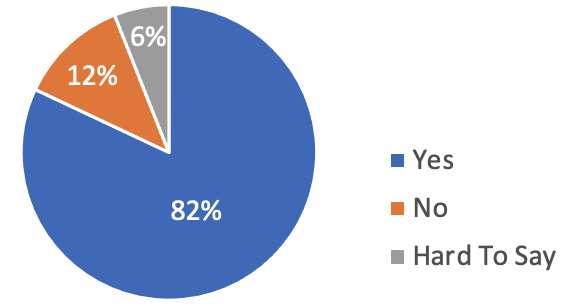}
\caption{Pie chart showing whether the generated sonnet be condensed into the planed keywords.}
\label{fig:pie}
\vspace{-1.7em}
\end{figure}

By comparing \VTwoName{} versus \ProsaicName{}, our human evaluation results already show that content planning contributes to discourse-level coherence. In addition, we provide the keywords along with the sonnet generated by \ModelName{}, and ask human annotators to judge if the sonnet can be condensed into those keywords. Results are shown in Figure \ref{fig:pie}. We observe  that 82\% of the time the planed keywords successfully guide the generation by providing salient points of the sonnet. We then conduct error analysis on the rest 18\%. Top two reasons among the fail cases are: 1)  the decoding step generates novel contents that are not represented by the keywords (8\%), and 2) the polishing step alters the original meaning of planed keywords (6\%).

\subsection{Limitation and Future Direction} 
Sonnets are divided in to multiple stanzas. Lines within a stanza are more interlaced than across stanza, and the start of a new one usually indicates transition to another viewpoint. Our current approach could not capture such structural characteristics during planning and generation, and we hope to investigate these features in future work. 

We also plan to extend this poem generation pipeline to other languages. For example, pretrained LMs (e.g. multilingual T5) and existing rhyming resources (r.g. \url{rhymes.woxikon.com} provides rhymes in 13 languages) already made the first and second component transferable to other languages. %Our current sentiment labels are coarse-grained and can not fully represent the nuances of human feelings. We plan to incorporate fine-grained sentiment planning in future work.

\section{Related Work}
\paragraph{Poetry Generation}
Automatic poetry generation before the deep learning age relies heavily on templates, norms, or rule-based approaches \cite{gervas2001expert, manurung2004evolutionary, manurung2012using}.
Neural approaches to automatic poetry generation pay little attention to the coherence issue of long poems. For example, \citet{wang2016chinese, lau2018deep, yi2018automatic, liu2019rhetorically} merely target at generating the first stanza (four lines) of a poem. For longer poems such as sonnets, \citet{ghazvininejad2016generating} propose to use related words as rhyme words to achieve topical relatedness, and later propose to generate discourse-level coherent English sonnets by French-English translation \cite{ghazvininejad2018neural}. \citet{van2020automatic} propose a naive RNN framework to generate rhyming verses from prosaic texts by imposing a priori word probability constraints. We on the other hand achieve discourse-level coherence by learning from standard, non-poetic texts. 

Other related works to boost the creativity of generated poems include adding rhetorical \cite{liu2019rhetorically} and influence factors (e.g., historical background) as latent variables \cite{yi2020mixpoet}.  To the best of our knowledge, we are the first to explore adding both figurative speeches and meter-and-rhyme constraints to poetry generation without relying on poetry data. 

\paragraph{Content Planning} Content planning for automatic text generation originates in the 1970s \cite{meehan1977tale}. 
Recently, the \textit{plan-and-write} generation framework has shown to be efficient in creative content generation \cite{wang2016chinese, martin2018event,peng2018towards,yao2019plan,gao2019discrete,goldfarb2019plan}. The framework employs a hierarchical paradigm and helps to produce more coherent and controllable generation than generating from scratch \cite{Fan2019StrategiesFS, goldfarb2020content}. However, all existing works under this line learn the storyline/plot from the target domain for improved coherence. We on the other hand adopt content planning to  disentangle the training from the decoding step which aims at circumventing the shortage of sizable creative contents for training supervised models. 
\section{Conclusion}
We investigate the possibility of generating sonnets without training on poems at all.
We propose a hierarchical planning-based framework to generate sonnets which first plans the high-level content of the poem, refine the predicted keywords by adding poetic features, and then achieve decoding-time control to impose the meter-and-rhyme constraints. Extensive automatic and expert evaluation show that our model can generate sonnets that use rich imagery and are globally coherent, poetic, and emotion provoking. 

\section*{Acknowledgments}
The authors would like to thank the members of PLUSLab and the anonymous reviewers for helpful comments. This work is supported in part by the DARPA Machine Common Sense (MCS) program under Cooperative Agreement N66001-19-2-4032. Yufei Tian is supported by an Amazon Fellowship.

% Entries for the entire Anthology, followed by custom entries
\bibliography{anthology,custom}
\bibliographystyle{acl_natbib}

\cleardoublepage

\end{document}